# Block-matching algorithm based on Harmony Search optimization for motion estimation

**Erik Cuevas**[1]
Departamento de Electrónica
Universidad de Guadalajara, CUCEI
Av. Revolución 1500, C.P 44430, Guadalajara, Jal, México
erik.cuevas@cucei.udg.mx

## Abstract

Motion estimation is one of the major problems in developing video coding applications. Among all motion estimation approaches, Block-matching (BM) algorithms are the most popular methods due to their effectiveness and simplicity for both software and hardware implementations. A BM approach assumes that the movement of pixels within a defined region of the current frame can be modeled as a translation of pixels contained in the previous frame. In this procedure, the motion vector is obtained by minimizing a certain matching metric that is produced for the current frame over a determined search window from the previous frame. Unfortunately, the evaluation of such matching measurement is computationally expensive and represents the most consuming operation in the BM process. Therefore, BM motion estimation can be viewed as an optimization problem whose goal is to find the best-matching block within a search space. The simplest available BM method is the Full Search Algorithm (FSA) which finds the most accurate motion vector through an exhaustive computation of all the elements of the search space. Recently, several fast BM algorithms have been proposed to reduce the search positions by calculating only a fixed subset of motion vectors despite lowering its accuracy. On the other hand, the Harmony Search (HS) algorithm is a population-based optimization method that is inspired by the music improvisation process in which a musician searches for harmony and continues to polish the pitches to obtain a better harmony. In this paper, a new BM algorithm that combines HS with a fitness approximation model is proposed. The approach uses motion vectors belonging to the search window as potential solutions. A fitness function evaluates the matching quality of each motion vector candidate. In order to save computational time, the approach incorporates a fitness calculation strategy to decide which motion vectors can be only estimated or actually evaluated. Guided by the values of such fitness calculation strategy, the set of motion vectors is evolved through HS operators until the best possible motion vector is identified. The proposed method has been compared to other BM algorithms in terms of velocity and coding quality. Experimental results demonstrate that the proposed algorithm exhibits the best balance between coding efficiency and computational complexity.

## 1. Introduction

Motion estimation plays important roles in a number of applications such as automobile navigation, video coding, surveillance cameras and so forth. The measurement of the motion vector is a fundamental problem in image processing and computer vision, which has been faced using several approaches [1-4]. The goal is to compute an approximation to the 2-D motion field - a projection of the 3-D velocities of surface points onto the imaging surface.

Video coding is currently utilized in a vast amount of applications ranging from fixed and mobile telephony, real-time video conferencing, DVD and high-definition digital television. Motion Estimation (ME) is an important part of any video coding system, since it can achieve significant compression by exploiting the temporal redundancy existing in a video sequence. Several ME methods have been studied aiming for a complexity reduction at video coding, such as block matching (BM) algorithms, parametric-based models [5], optical flow [6] and pel-recursive techniques [7]. Among such methods, BM seems to be the most popular technique due to its effectiveness and simplicity for both software and hardware implementations [8]. Furthermore, in order to reduce the computational complexity in ME, many BM algorithms have been proposed and used in implementations of various video compression standards such as MPEG-4 [9] and H.264 [10].

---

[1] Tel +52 33 1378 5900,  Ext. 27714, E-mail: erik.cuevas@cucei.udg.mx

In BM algorithms, the video frames are partitioned in non-overlapping blocks of pixels. Each block is predicted from a block of equal size in the previous frame. Specifically, for each block in the current frame, we search for a best matching block within a searching window in the previous frame that minimizes a certain matching metric. The most used matching measure is the Sum of Absolute Differences (SAD) which is computationally expensive and represents the most consuming operation in the BM process. The best matching block found represents the predicted block, whose displacement from the previous block is represented by a transitional motion vector (MV). Therefore, BM is essentially an optimization problem, with the goal of finding the best matching block within a search space.

The full search algorithm (FSA) [11] is the simplest block-matching algorithm that can deliver the optimal estimation solution regarding a minimal matching error as it checks all candidates one at a time. However, such exhaustive search and full-matching error calculation at each checking point yields an extremely computational expensive BM method that seriously constraints real-time video applications.

In order to decrease the computational complexity of the BM process, several BM algorithms have been proposed considering the following three techniques: (1) using a fixed pattern: which means that the search operation is conducted over a fixed subset of the total search window. The Three Step Search (TSS) [12], the New Three Step Search (NTSS) [13], the Simple and Efficient TSS (SES) [14], the Four Step Search (4SS) [15] and the Diamond Search (DS) [16] are some of its well-known examples. Although such approaches have been algorithmically considered as the fastest, they are not able eventually to match the dynamic motion-content, delivering false motion vectors (image distortions). (2) Reducing the search points: in this method, the algorithm chooses as search points exclusively those locations which iteratively minimize the error-function (SAD values). This category includes: the Adaptive Rood Pattern Search (ARPS) [17], the Fast Block Matching Using Prediction (FBMAUPR) [18], the Block-based Gradient Descent Search (BBGD) [19] and the Neighbourhood Elimination algorithm (NE) [20]. Such approaches assume that the error-function behaves monotonically, holding well for slow-moving sequences; however, such properties do not hold true for other kind of movements in video sequences [21], which risks on algorithms getting trapped into local minima. (3) Decreasing the computational overhead for every search point, which means the matching cost (SAD operation) is replaced by a partial or a simplify version that features less complexity. The New pixel-Decimation (ND) [22], the Efficient Block Matching Using Multilevel Intra and Inter-Sub-blocks [13] and the Successive Elimination Algorithm [23], all assume that all pixels within each block move by the same amount and a good estimate of the motion could be obtained through only a fraction of the pixel pool. However, since only a fraction of pixels enters into the matching computation, the use of these regular sub-sampling techniques can seriously affect the accuracy of the detection of motion vectors due to noise or illumination changes.

Another popular group of BM algorithms employ spatiotemporal correlation, using the neighboring blocks in spatial and temporal domain. The main advantage of these algorithms is that they alleviate the local minimum problem to some extent. Since the new initial or predicted search center is usually closer to the global minimum, the chance of getting trapped in a local minimum decreases. This idea has been incorporated by many fast block motion estimation algorithms such as the enhanced predictive zonal search (EPZS) [24] and the UMHexagonS [25]. However, the information delivered by the neighboring blocks occasionally conduces to false initial search points, producing distorted motion vectors. Such problem is typically caused when very small objects moves during the image sequence [26].

Alternatively, evolutionary approaches such as genetic algorithms (GA) [27] and particle swarm optimization (PSO) [28] are well known for locating the global optimum in complex optimization problems. Despite of such fact, only few evolutionary approaches have specifically addressed the problem of BM, such as the light-weight genetic block matching (LWG) [29], the genetic four-step search (GFSS) [30] and the PSO-BM [31].

Although these methods support an accurate identification of the motion vector, their spending times are very long in comparison to other BM techniques.

On the other hand, the Harmony Search (HS) algorithm introduced by Geem, Kim, and Loganathan [32] is one of the population-based evolutionary heuristics algorithms which are based on the metaphor of the improvisation process that occurs when a musician searches for a better state of harmony. The HS generates a new candidate solution from all existing solutions. In HS, the solution vector is analogous to the harmony in music, and the local and global search schemes are analogous to musician's improvisations. In comparison to other meta-heuristics in the literature, HS imposes fewer mathematical requirements as it can be easily adapted for solving several sorts of engineering optimization challenges [33,34]. Furthermore, numerical comparisons have demonstrated that the evolution for the HS is faster than GA [33,35,36], attracting ever more attention. It has been successfully applied to solve a wide range of practical optimization problems such as structural optimization, parameter estimation of the nonlinear Muskingum model, design optimization of water distribution networks, vehicle routing, combined heat and power economic dispatch, design of steel frames, bandwidth-delay-constrained least-cost multicast routing, computer vision, among others [35–43].

A main difficulty applying HS to solve real-world problems is that it usually needs a large number of fitness evaluations before an acceptable result can be obtained. In practice, however, fitness evaluations are not always straightforward because either an explicit fitness function does not exist (an experiment is needed instead) or the evaluation of the fitness function is computationally demanding. Furthermore, since random numbers are involved in the calculation of new individuals, they may encounter the same positions (repetition) that other individuals have visited in previous iterations, especially when the individuals are confined to a small area.

The problem of considering expensive fitness evaluations has already been faced in the field of evolutionary algorithms (EA) and is better known as fitness approximation [44]. In such approach, the idea is to estimate the fitness value of so many individuals as it is possible instead of evaluating the complete set. Such estimations are based on an approximate model of the fitness landscape. Thus, the individuals to be evaluated and those to be estimated are determined following some fixed criteria which depend on the specific properties of the approximate model [45]. The models involved at the estimation can be built during the actual EA run, since EA repeatedly sample the search space at different points [46]. There are many possible approximation models and several have already been used in combination with EA (e.g. polynomials [47], the kriging model [48], the feed-forward neural networks that includes multi-layer Perceptrons [49] and radial basis-function networks [50]). These models can be either global, which make use of all available data or local which make use of only a small set of data around the point where the function is to be approximated. Local models, however, have a number of advantages [46]: they are well-known and suitably established techniques with relatively fast speeds. Moreover, they consider the intuitively most important information: the closest neighbors.

In this paper, a new BM algorithm that combines HS with a fitness approximation model is proposed. Since the proposed method approaches the BM process as an optimization problem, its overall operation can be formulated as follows: First, a population of individuals is initialized where each individual represents a motion vector candidate (a search location). Then, the set of HS operators is applied at each iteration in order to generate a new population. The procedure is repeated until convergence is reached whereas the best solution is expected to represent the most accurate motion vector. In the optimization process, the quality of each individual is evaluated through a fitness function which represents the SAD value corresponding to each motion vector. In order to save computational time, the approach incorporates a fitness estimation strategy to decide which search locations can be only estimated or actually evaluated. The proposed method has been compared to other BM algorithms in terms of velocity and coding quality. Experimental results show that the proposed BM algorithm exhibits the best trade-off between coding efficiency and computational complexity.

The overall paper is organized as follows: Section 2 holds a brief description about the HS method In Section 3, the fitness calculation strategy for solving the expensive optimization problem is presented. Section 4 provides background about the BM motion estimation issue while Section 5 exposes the final BM algorithm as a combination of HS and the fitness calculation strategy. Section 6 demonstrates experimental results for the proposed approach over standard test sequences and some conclusions are drawn in Section 7.

## 2. Harmony search algorithm

### 2.1. The Harmony Search Algorithm

In the basic HS, each solution is called a ''harmony'' and is represented by an *n*-dimension real vector. An initial population of harmony vectors are randomly generated and stored within a Harmony Memory (HM). A new candidate harmony is thus generated from the elements in the HM by using a memory consideration operation either by a random re-initialization or a pitch adjustment operation. Finally, the HM is updated by comparing the new candidate harmony and the worst harmony vector in the HM. The worst harmony vector is replaced by the new candidate vector in case it is better than the worst harmony vector in the HM. The above process is repeated until a certain termination criterion is met. The basic HS algorithm consists of three basic phases: HM initialization, improvisation of new harmony vectors and updating of the HM. The following discussion addresses details about each stage.

#### 2.1.1. Initializing the problem and algorithm parameters

In general, the global optimization problem can be summarized as follows: min $f(\mathbf{x})$ : $x(j) \in [l(j), u(j)]$, $j = 1, 2, \ldots, n$, where $f(\mathbf{x})$ is the objective function, $\mathbf{x} = (x(1), x(2), \ldots, x(n))$ is the set of design variables, $n$ is the number of design variables, and $l(j)$ and $u(j)$ are the lower and upper bounds for the design variable $x(j)$, respectively. The parameters for HS are the harmony memory size, i.e., the number of solution vectors lying on the harmony memory (HM), the harmony-memory consideration rate (*HMCR*), the pitch adjusting rate (*PAR*), the distance bandwidth (*BW*) and the number of improvisations (*NI*) which represents the total number of iterations. It is obvious that the performance of HS is strongly influenced by parameter values which determine its behavior.

#### 2.1.2. Harmony memory initialization

In this stage, initial vector components at HM, i.e., *HMS* vectors, are configured. Let $\mathbf{x}_i = \{x_i(1), x_i(2), \ldots, x_i(n)\}$ represent the *i*-th randomly-generated harmony vector: $x_i(j) = l(j) + (u(j) - l(j)) \cdot \text{rand}(0,1)$ for $j = 1, 2, \ldots, n$ and $i = 1, 2, \ldots, HMS$, where rand(0,1) is a uniform random number between 0 and 1. Then, the HM matrix is filled with the *HMS* harmony vectors as follows:

$$\text{HM} = \begin{bmatrix} \mathbf{x}_1 \\ \mathbf{x}_2 \\ \vdots \\ \mathbf{x}_{HMS} \end{bmatrix} \quad (1)$$

#### 2.1.3. Improvisation of new harmony vectors

In this phase, a new harmony vector $\mathbf{x}_{new}$ is built by applying the following three operators: memory consideration, random re-initialization and pitch adjustment. Generating a new harmony is known as 'improvisation'. In the memory consideration step, the value of the first decision variable $x_{new}(1)$ for the new vector is chosen randomly from any of the values already existing in the current HM i.e., from the set $\{x_1(1), x_2(1), \ldots, x_{HMS}(1)\}$. For this operation, a uniform random number $r_1$ is generated within the range [0, 1]. If $r_1$ is less than *HMCR*, the decision variable $x_{new}(1)$ is generated through memory considerations; otherwise, $x_{new}(1)$ is obtained from a random re-initialization between the search bounds $[l(1), u(1)]$. Values of the other decision variables $x_{new}(2), x_{new}(3), \ldots, x_{new}(n)$ are also chosen accordingly. Therefore, both operations, memory consideration and random re-initialization, can be modelled as follows:

$$x_{new}(j) = \begin{cases} x_i(j) \in \{x_1(j), x_2(j), \ldots, x_{HMS}(j)\} & \text{with probability } HMCR \\ l(j) + (u(j) - l(j)) \cdot \text{rand}(0,1) & \text{with probability } 1\text{-}HMCR \end{cases} \quad (2)$$

Every component obtained by memory consideration is further examined to determine whether it should be pitch-adjusted. For this operation, the Pitch-Adjusting Rate (*PAR*) is defined as to assign the frequency of the adjustment and the Bandwidth factor (*BW*) to control the local search around the selected elements of the HM. Hence, the pitch adjusting decision is calculated as follows:

$$x_{new}(j) = \begin{cases} x_{new}(j) = x_{new}(j) \pm \text{rand}(0,1) \cdot BW & \text{with probability } PAR \\ x_{new}(j) & \text{with probability } (1\text{-}PAR) \end{cases} \quad (3)$$

Pitch adjusting is responsible for generating new potential harmonies by slightly modifying original variable positions. Such operation can be considered similar to the mutation process in evolutionary algorithms. Therefore, the decision variable is either perturbed by a random number between 0 and *BW* or left unaltered. In order to protect the pitch adjusting operation, it is important to assure that points lying outside the feasible range $[l, u]$ must be re-assigned i.e., truncated to the maximum or minimum value of the interval.

*2.1.4. Updating the harmony memory*

After a new harmony vector $x_{new}$ is generated, the harmony memory is updated by the survival of the fit competition between $x_{new}$ and the worst harmony vector $x_w$ in the HM. Therefore $x_{new}$ will replace $x_w$ and become a new member of the HM in case the fitness value of $x_{new}$ is better than the fitness value of $x_w$.

*2.2. Computational procedure*

The computational procedure of the basic HS can be summarized as follows [18]:

Step 1: Set the parameters *HMS*, *HMCR*, *PAR*, *BW* and *NI*.
Step 2: Initialize the HM and calculate the objective function value of each harmony vector.
Step 3: Improvise a new harmony $\mathbf{x}_{new}$ as follows:
    for ($j = 1$ to $n$) do
      if ($r_1 < HMCR$) then
        $x_{new}(j) = x_a(j)$ where $a$ is element of $(1, 2, \ldots, HMS)$ randomly selected
        if ($r_2 < PAR$) then
          $x_{new}(j) = x_{new}(j) \pm r_3 \cdot BW$ where $r_1, r_2, r_3 \in \text{rand}(0,1)$
        end if
        if $x_{new}(j) < l(j)$
          $x_{new}(j) = l(j)$
        end if
        if $x_{new}(j) > u(j)$
          $x_{new}(j) = u(j)$
        end if
      else
        $x_{new}(j) = l(j) + r \cdot (u(j) - l(j))$, where $r \in \text{rand}(0,1)$
      end if
    end for
Step 4: Update the *HM* as $\mathbf{x}_w = \mathbf{x}_{new}$ if $f(\mathbf{x}_{new}) < f(\mathbf{x}_w)$
Step 5: If *NI* is completed, the best harmony vector $\mathbf{x}_b$ in the HM is returned; otherwise go back to step 3.
## 3. Fitness approximation method

Evolutionary algorithms that use fitness approximation aim to find the global minimum of a given function considering only a very few number of function evaluations and a large number of estimations, based on an approximate model of the function landscape. In order to apply such approach, it is necessary that the objective function implicates a very expensive evaluation and consists of few dimensions (up to five) [51]. Recently, several fitness estimators have been reported in the literature [47-50] in which the number of function evaluations is considerably reduced to hundreds, dozens, or even less. However, most of these methods produce complex algorithms whose performance is conditioned to the quality of the training phase and the learning algorithm in the construction of the approximation model.

In this paper, we explore the use of a local approximation scheme based on the nearest-neighbor-interpolation (NNI) for reducing the function evaluation number. The model estimates the fitness values based on previously evaluated neighboring individuals which have been stored during the evolution process. At each generation, some individuals of the population are evaluated through the accurate (real) fitness function while the other remaining individuals are only estimated. The positions to be accurately evaluated are determined based on their proximity to the best individual or regarding their uncertain fitness value.

*3.1 Updating the individual database*

In a fitness approximation method, every evaluation of an individual produces one data point (individual position and fitness value) that is potentially taken into account for building the approximation model during the evolution process. Therefore, in our proposed approach, we keep all seen-so-far evaluated individuals and their respective fitness values within a history array **T** which is employed to select the closest neighbor and to estimate the fitness value of a new individual. Thus, each element of **T** consists of two parts: the individual position and its respective fitness value. The array **T** begins with null elements in the first iteration. Then, as the optimization process evolves, new elements are added. Since the goal of a fitness approximation approach is to evaluate the least possible number of individuals, only few elements are contained in **T**.

*3.2 Fitness calculation strategy*

This section explains the strategy to decide which individuals are to be evaluated or estimated. The proposed fitness calculation scheme estimates most of fitness values to reduce the computational overhead at each generation. In the model, those individuals positioned nearby the individual with the best fitness value at the array **T** (Rule 1) are evaluated by using the actual fitness function. Such individuals are important as they possess a stronger influence over the evolution process than the others. Moreover, it also evaluates those individuals placed in regions of the search space with no previous evaluations (Rule 2). Fitness values for these individuals are uncertain since there is no close reference (close points contained in **T**) to calculate their estimates.

The remaining individuals, for which there exist a close point that is previously evaluated and its fitness value is not the best contained in the array **T**, are estimated using the NNI criterion (Rule 3). Thus, the fitness value of an individual is approximated by assigning the same fitness value that the nearest individual stored in **T**.

Therefore, the fitness computation model follows three important rules to evaluate or estimate fitness values:

1. *Exploitation rule (evaluation).* If a new individual (search position) $P$ is located closer than a distance $d$ with respect to the nearest individual $L_q$ contained in **T** ( $q = 1, 2, 3, \ldots, m$; where $m$ is the number of elements contained in **T**), whose fitness value $F_{L_q}$ corresponds to the best fitness value, then the fitness value of $P$ is evaluated by using the actual fitness function. Figure 1a draws the rule procedure.

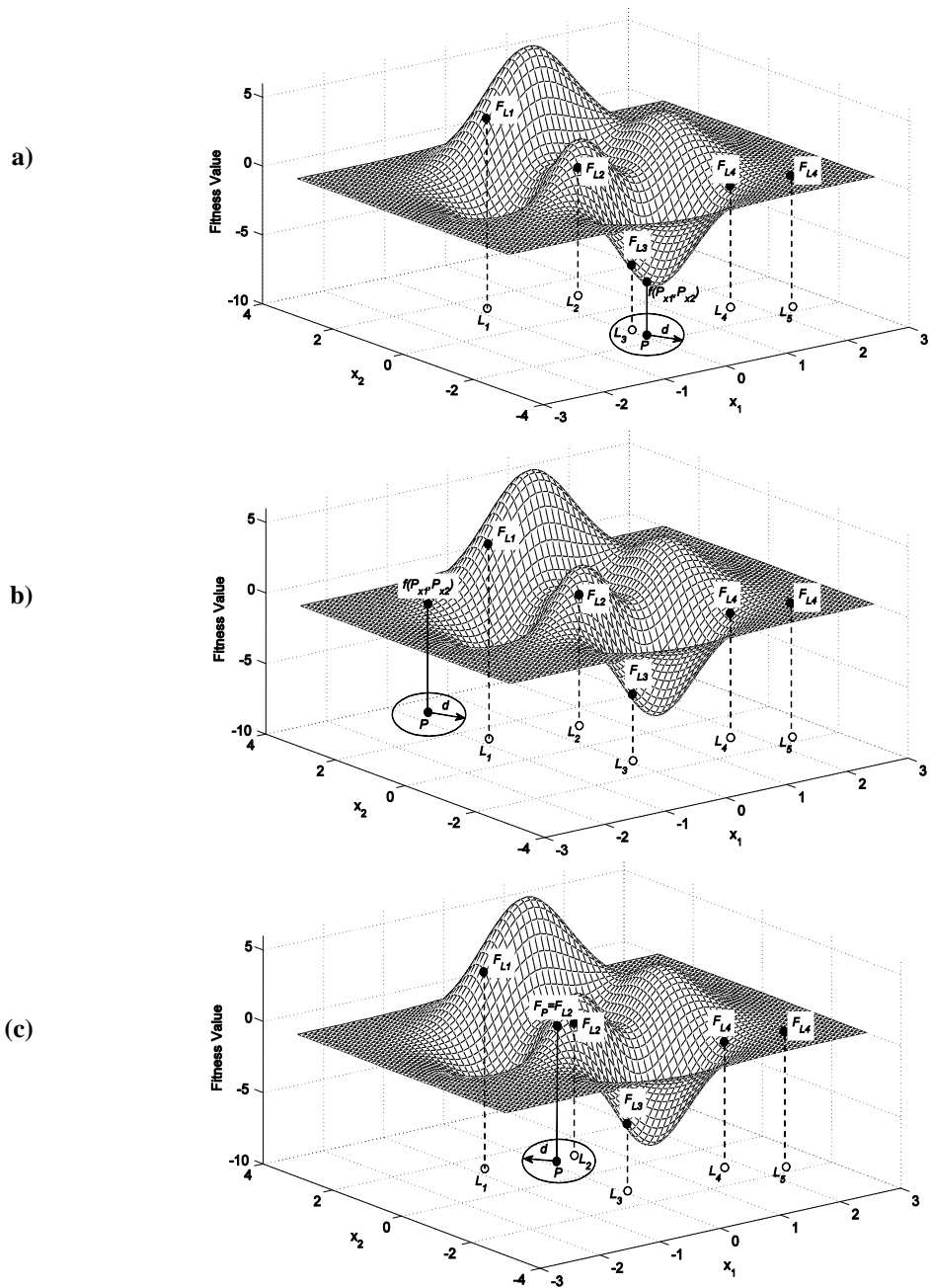

**Fig. 1.** The fitness calculation strategy. (a) According to the rule 1, the individual (search position) *P* is evaluated since it is located closer than a distance *d* with respect to the nearest individual location $L_3$. Therefore, the fitness value $F_{L_3}$ corresponds to the best fitness value (minimum). (b) According to the rule 2, the search point *P* is evaluated and there is no close reference within its neighborhood. (c) According to rule 3, the fitness value of *P* is estimated by means of the NNI-estimator, assigning $F_P = F_{L_2}$

2. *Exploration rule (evaluation).* If a new individual *P* is located further away than a distance *d* with respect to the nearest individual $L_q$ contained in **T,** then its fitness value is evaluated by using the actual fitness function. Figure 1b outlines the rule procedure.

3. *NNI rule (estimation)*. If a new individual $P$ is located closer than a distance $d$ with respect to the nearest individual $L_q$ contained in **T**, whose fitness value $F_{L_q}$ does not correspond to the best fitness value, then its fitness value is estimated assigning it the same fitness that $L_q$ ($F_P = F_{L_q}$). Figure 1c sketches the rule procedure.

The $d$ value controls the trade-off between the evaluation and the estimation of search locations. Typical values of $d$ range from 1 to 4. Values close to 1 improve the precision at the expense of a higher number of fitness evaluations (the number of evaluated individuals is more than the number of estimated). On the other hand, values close to 4 decrease the computational complexity at the price of poor accuracy (decreasing the number of evaluation and increasing the number of estimations). After exhaustive experimentation, it has been determined that a value of $d=3$ represents the best trade-off between computational overhead and accuracy, so it is used throughout the study. The proposed method, from an optimization perspective, favors the exploitation and exploration in the search process. For the exploration, the method evaluates the fitness function of new search locations which have been located far from previously calculated positions. Additionally, it also estimates those that are closer. For the exploitation, the proposed method evaluates the fitness function of those new searching locations which are placed nearby the position that holds the minimum fitness value seen so far. Such fact is considered as an strong evidence that the new location could improve the "best value" (the minimum) already found.

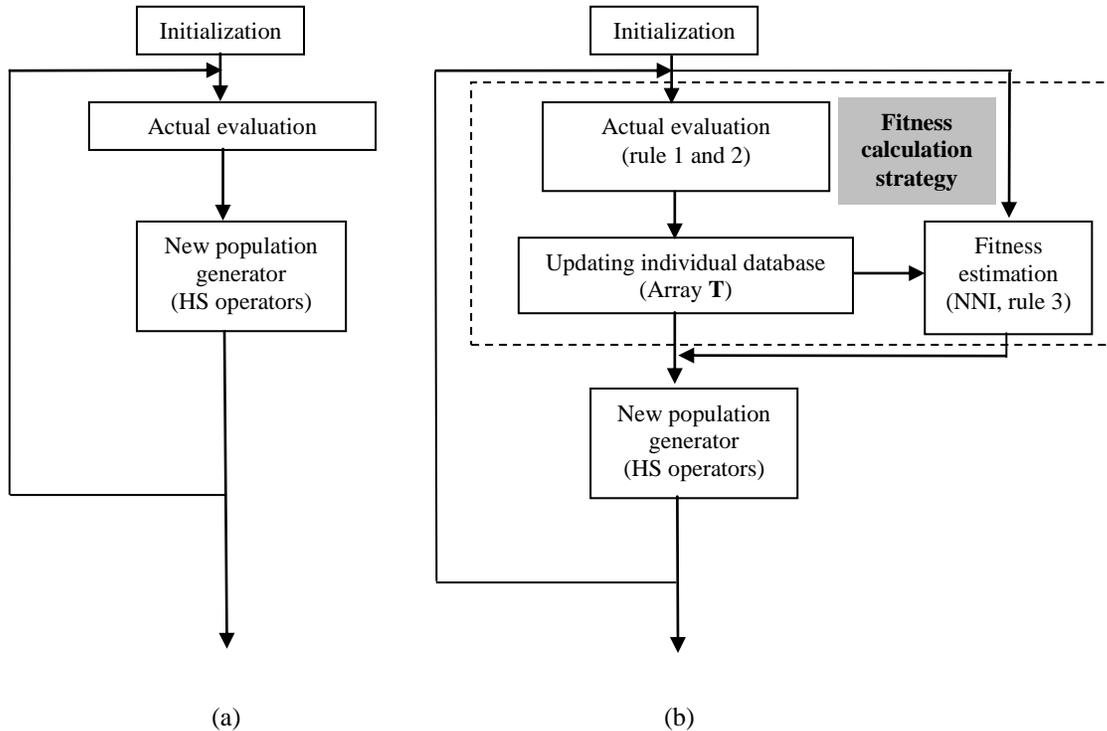

(a)            (b)

**Fig. 2.** Differences between the conventional HS and the proposed HS optimization method. (a) Conventional HS and (b) the HS algorithm including the fitness calculation strategy

In order to know which rule must be applied by the fitness approximation strategy, considering a new search position $P$, it is only necessary to identify the closest individual $L_q$ which is contained in T. Then, it is inquired if the positional relationship between them and the fitness value $F_{L_q}$ of $L_q$ fulfill the properties imposed by each rule (distance, if $F_{L_q}$ is the best fitness values contained in **T**, etc). As the number of elements of the array **T** is very limited, the computational complexity resulting from such operations is negligible. Fig. 1 illustrates the procedure of fitness computation for a new solution (point $P$). In the problem, the objective function *f* is minimized with respect to two parameters ($x_1, x_2$). In all figures (Figs. 1(a), (b) and (c)), the individual database array **T** contains five different elements ($L_1, L_2, L_3, L_4, L_5$) with their corresponding fitness values ($F_{L_1}, F_{L_2}, F_{L_3}, F_{L_4}, F_{L_5}$). Figures 1(a) and (b) show the fitness evaluation ($f(x_1, x_2)$) of the new solution *P*, following the rule 1 and 2 respectively, whereas Fig. 1(c) present the fitness estimation of *P* using the NNI approach which is laid by rule 3.

*3.3 Proposed HS optimization method*

The coupling of HS and the fitness approximation strategy is presented in this paper as an optimization approach. The only difference between the conventional HS and the enhanced HS method is the fitness calculation scheme. In the proposed algorithm, only some individuals are actually evaluated (Rules 1 and 2) at each generation. All other fitness values for the rest are estimated using the NNI-approach (Rule 3). The estimation is executed by using the individuals previously calculated which are contained in the array **T**.

Fig. 2 shows the difference between the conventional HS and the proposed version. It is clear that two new blocks have been added, the fitness estimation and the updating individual database. Both elements and the actual evolution block, represent the fitness calculation strategy just as it has been explained at Section 3.2. As a result, the HS approach can substantially reduce the number of function evaluations preserving the good search capabilities of HS.

**4. Motion estimation and block matching**

For motion estimation, in a BM algorithm, the current frame of an image sequence $I_t$ is divided into non-overlapping blocks of *N*x*N* pixels. For each template block in the current frame, the best matched block within a search window (*S*) of size (2*W*+1)x(2*W*+1) in the previous frame $I_{t-1}$ is determined, where *W* is the maximum allowed displacement. The position difference between a template block in the current frame and the best matched block in the previous frame is called the Motion Vector (MV) (see Fig. 3). Therefore, BM can be viewed as an optimization problem, with the goal of finding the best MV within a search space.

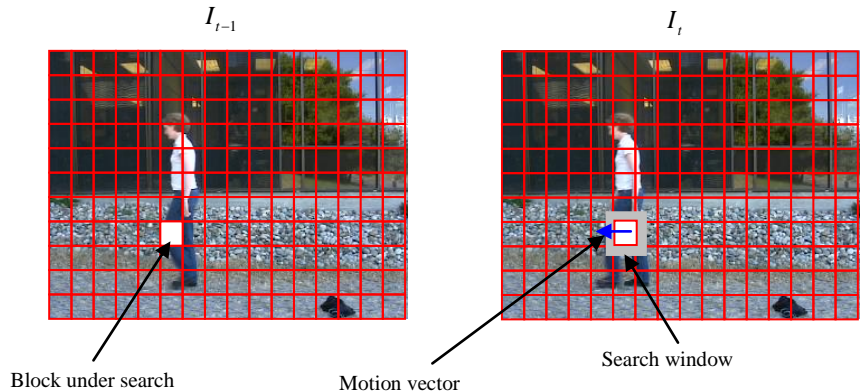

**Fig. 3.** Block Matching procedure.

The most well-known matching criterion for BM algorithms is the sum of absolute difference (SAD). It is defined in Eq. (5) considering a template block at position (*x*, *y*) in the current frame and the candidate block at position $(x+\hat{u}, y+\hat{v})$ in the previous frame $I_{t-1}$.

$$\text{SAD}(\hat{u},\hat{v}) = \sum_{j=0}^{N-1}\sum_{i=0}^{N-1} \left| g_t(x+i, y+j) - g_{t-1}(x+\hat{u}+i, y+\hat{v}+j) \right| \qquad (4)$$

where $g_t(\cdot)$ is the gray value of a pixel in the current frame $I_t$ and $g_{t-1}(\cdot)$ is the gray level of a pixel in the previous frame $I_{t-1}$. Therefore, the MV in $(u,v)$ is defined as follows:

$$(u,v) = \arg \min_{(u,v) \in S} \text{SAD}(\hat{u},\hat{v}) \qquad (5)$$

where $S = \{(\hat{u},\hat{v}) | -W \leq \hat{u},\hat{v} \leq W \text{ and } (x+\hat{u}, y+\hat{v}) \text{ is a valid pixel position } I_{t-1}\}$. As it can be seen, the computing of such matching criterion is a consuming time operation which represents the bottle-neck in the BM process.

In the context of BM algorithms, the FSA is the most robust and accurate method to find the MV. It tests all possible candidate blocks from $I_{t-1}$ within the search area to find the block with minimum SAD. For the maximum displacement of $W$, the FSA requires $(2W+1)^2$ search points. For instance, if the maximum displacement $W$ is ±7, the total search points are 225. Each SAD calculation requires $2N^2$ additions and the total number of additions for the FSA to match a 16×16 block is 130,560. Such computational requirement makes the application of FSA difficult for real time applications.

## 5. BM algorithm based on HS with the estimation strategy

FSA finds the global minimum (the accurate MV), considering all locations within the search space $S$. Nevertheless, the approach has a high computational cost for practical use. In order to overcome such a problem, many fast algorithms have been developed yielding only a poorer precision than the FSA. A better BM algorithm should spend less computational time on searching and obtaining accurate motion vectors (MVs).

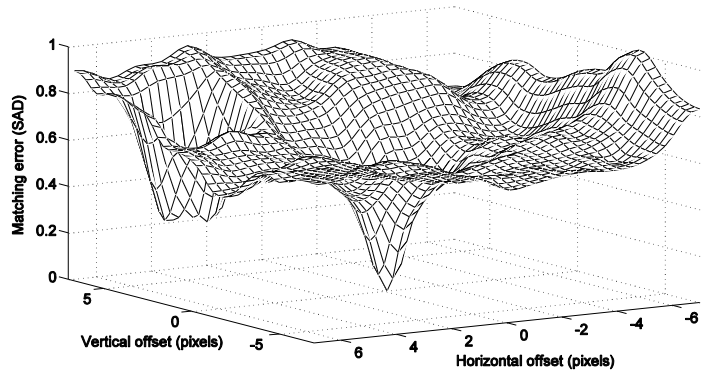

**Fig. 4.** Common non-uni-modal error surface with multiple local minimum error points

The BM algorithm proposed at this paper is comparable to the fastest algorithms and delivers a similar precision to the FSA approach. Since most of fast algorithms use a regular search pattern or assume a characteristic error function (uni-modal) for searching the motion vector, they may get trapped into local minima considering that for many cases (i.e., complex motion sequences) an uni-modal error is no longer valid. Fig. 4 shows a typical error surface (SAD values) which has been computed around the search window for a fast-moving sequence. On the other hand, the proposed BM algorithm uses a non-uniform search pattern for locating global minimum distortion. Under the effect of the HS operators, the search locations vary from generation to generation, avoiding to get trapped into a local minimum. Besides, since the proposed algorithm uses a fitness calculation strategy for reducing the evaluation of the SAD values, it requires fewer search positions.

In the algorithm, the search space *S* consists of a set of 2-D motion vectors $\hat{u}$ and $\hat{v}$ representing the *x* and *y* components of the motion vector, respectively. The particle is defined as:

$$P_i = \{\hat{u}_i, \hat{v}_i \mid -W \leq \hat{u}_i, \hat{v}_i \leq W\} \tag{6}$$

where each particle *i* represents a possible motion vector. In this paper, the search windows, considered in the simulations, are set to ±8 and ±16 pixels. Both configurations are selected in order to compare the results with other approaches presented in the literature.

*5.1 Initial population*

The first step in HS optimization is to generate an initial group of individuals. The standard literature of evolutionary algorithms generally suggests the use of random solutions as the initial population, considering the absence of knowledge about the problem [52]. However, several studies [53-56] have demonstrated that the use of solutions generated through some domain knowledge to set the initial population (i.e., non-random solutions) can significantly improve its performance. In order to obtain appropriate initial solutions (based on knowledge), an analysis over the motion vector distribution was conducted. After considering several sequences (see Table 1 and Fig. 9), it can be seen that 98% of the MVs are found to lie at the origin of the search window for a slow-moving sequence such as the one at *Container*, whereas complex motion sequences, such as the *Carphone* and the *Foreman* examples, have only 53.5% and 46.7% of their MVs in the central search region. The *Stefan* sequence, showing the most complex motion content, has only 36.9%. Figure 5 shows the surface of the MV distribution for the *Foreman* and the *Stefan*. On the other hand, although it is less evident, the MV distribution of several sequences shows small peaks at some locations lying away from the center as they are contained inside a rectangle that is shown in Fig. 5(b) and 5(d) by a white overlay. Real-world moving sequences concentrate most of the MVs under a limit due to the motion continuity principle [16]. Therefore, in this paper, initial solutions are selected from five fixed locations which represent points showing the higher concentration in the MV distribution, just as it is shown by Figure 6.

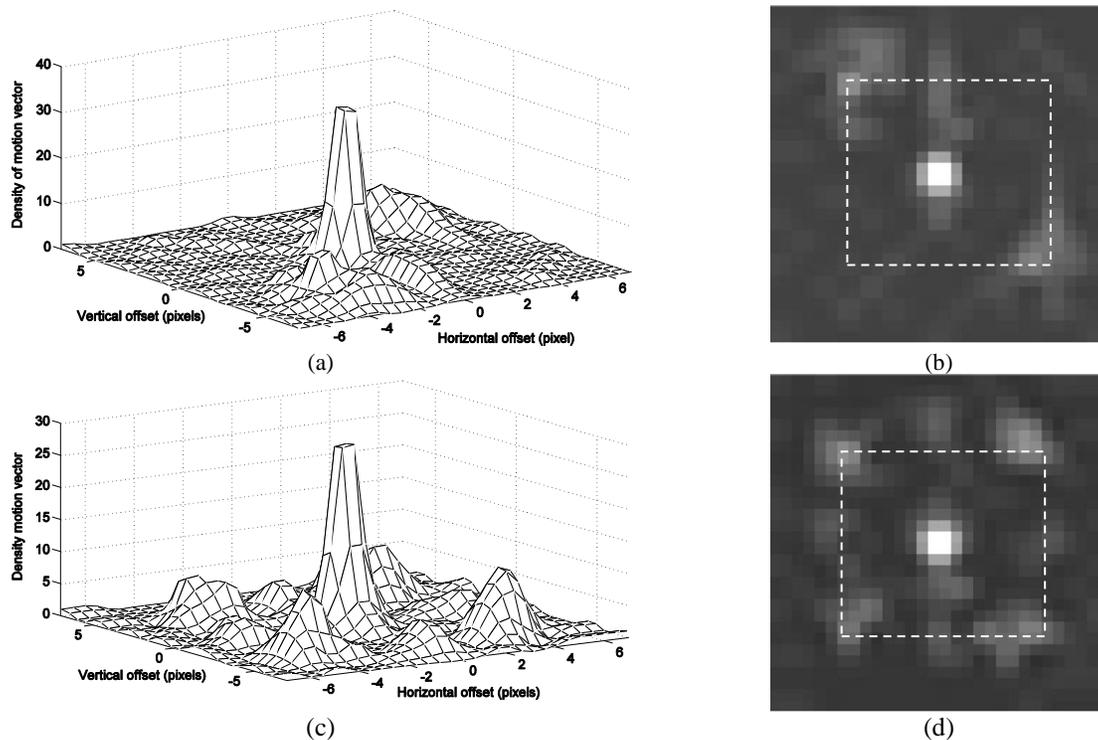

**Fig. 5.** Motion vector distribution for *Foreman* and Stefan sequences. (a)-(b) MV distribution for the *Foreman* sequence. (c)-(d) MV distribution for the *Stefan* sequence.

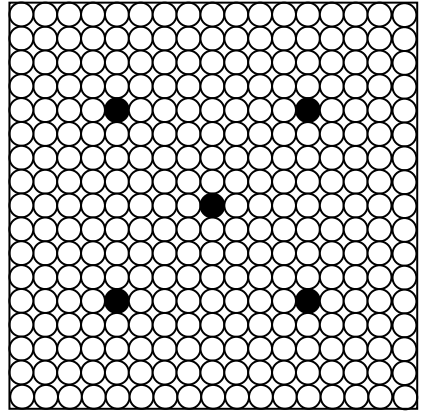

**Fig. 6.** Fixed pattern of five elements in the search window of ±8, used as initial solutions.

Since most movements suggest displacements near to the center of the search window [12,13], the initial solutions shown by Fig. 6 are used as initial position for the HS algorithm. This consideration is taken regardless of the search window employed (±8 or ±16).

*5.2 Tuning of the HS algorithm*

The performance of HS is strongly influenced by parameter values which determine its behavior. HS incorporates several parameters such as the population size, the operator's probabilities (as *HMCR* and *PAR*) or the total number of iterations (*NI*). Determining the most appropriate parameter values for a determined problem is a complex issue, since such parameters interact to each other in a highly nonlinear manner and there are not mathematical models of such interaction. Throughout the years, two main types of methods have been proposed for setting up parameter values of an evolutionary algorithm: off-line and on-line strategies [57]. An off-line method (called tuning) searches for the best set of parameter values through experimentation. Once defined, these values remain fixed. Such methodology is appropriate when the optimization problem maintains the same properties (dimensionality, multimodality, unconstrained, etc) each time that the EA is applied. On the other hand, on-line methods focus on changing parameter values during the search process of the algorithm. Thus, the strategy must decide when to change parameter values and determine new values. Therefore, these methods are indicated when EA faces optimizations problems with dimensional variations or restriction changes, etc.

Considering that the optimization problem outlined by the BM process maintains the same properties (same dimensions and similar error landscapes), the off-line method has been used for tuning the HS algorithm. Therefore, after exhaustive experimentation, the following parameters have been found as the best parameter set, *HMCR*=0.7, *PAR*=0.3. Considering that the proposed approach is tested by using two different search windows (±8 and ±16), the values of *BW* and *NI* have different configurations depending on the selected search window. Therefore, it is employed *BW*=8 and *NI*=25 in the case of a window search of ±8 whereas the case of ±16, it uses *BW*=16 and *NI*=45. Once such configurations are defined, the parameter set is kept for all test sequences through all experiments.

*5.3 The HS-BM algorithm*

The goal of our BM-approach is to reduce the number of evaluations of the SAD values (real fitness function) avoiding any performance loss and achieving an acceptable solution. The HS-BM method is listed below:

**Step 1:**      Set the HS parameters. *HMCR*=0.7, *PAR*=0.3, *BW*=8 in case of a search window of ±8 and 16 in case of ±16.

**Step 2:** Initialize the harmony memory with five individuals (*HMS*=5), where each decision variable $u$ and $v$ of the candidate motion vector $\text{MV}_a$ is set according to the fixed pattern shown in Fig. 6. Considering $a \in (1, 2, \ldots, HMS)$. Define also the individual database array **T**, as an empty array.

**Step 3:** Compute the fitness values for each individual according to the fitness calculation strategy presented in Section 3. Since all individuals of the initial population fulfil rule 2 conditions, they are evaluated through a real fitness function (calculating the real SAD values).

**Step 4:** Update the new evaluations in the individual database array **T**.

**Step 5:** Determine the candidate solution $\text{MV}_w$ of *HMS* holding the worst fitness value.

**Step 6:** improvise a new harmony $\text{MV}_{new}$ such that:
for ($j$ = 1 to 2) do
  if ($r_1 < HMCR$) then
    $\text{MV}_{new}(j) = \text{MV}_a(j)$ where $a$ is element of $(1, 2, \ldots, HMS)$ randomly selected
    if ($r_2 < PAR$) then
      $\text{MV}_{new}(j) = \text{MV}_{new}(j) \pm r_3 \cdot BW$ where $r_1, r_2, r_3 \in (0,1)$
      if $\text{MV}_{new}(j) < l(j)$
      $\text{MV}_{new}(j) = l(j)$
      end if
      if $\text{MV}_{new}(j) > u(j)$
      $\text{MV}_{new}(j) = u(j)$
      end if
    end if
  else
    $\text{MV}_{new}(j) = 1 + \text{round}(r \cdot E_p)$, where $r \in (-1,1)$, $E_p = 8$ or 16.
  end if
end for

$E_p = 8$, in case of a search window of $\pm 8$ and 16 in case of $\pm 16$.

**Step 7:** Compute the fitness value of $\text{MV}_{new}$ by using the fitness calculation strategy presented in Section 3.

**Step 8:** Update the new evaluation in the individual database array **T**.

**Step 9:** Update *HM*. In case that the fitness value (evaluated or approximated) of the new solution $\text{MV}_{new}$, is better than the solution $\text{MV}_w$, such position is selected as an element of *HM*, otherwise the solution $\text{MV}_w$ remains.

**Step 10:** Determine the best individual of the current new population. If the new fitness (SAD) value is better than the old best fitness value, then update $\hat{u}_{best}$, $\hat{v}_{best}$.

**Step 11:** If the number of iterations (*NI*) has been reached (25 in the case of a search window of $\pm 8$ and 45 for $\pm 16$), then the MV is $\hat{u}_{best}$, $\hat{v}_{best}$; otherwise go back to Step 5.

Thus, the proposed HS-BM algorithm considers different search locations, 30 in the case of a search window of ±8 and 50 for ±16, during the complete optimization process (which consists of 25 and 45 different iterations depending on the search window, plus the five initial positions). However, only a few search locations are evaluated using the actual fitness function (5-14, in the case of a search window of ±8 and 7-22, for ±16) while the remaining positions are just estimated. Therefore, as the evaluated individuals and their respective fitness values are exclusively stored in the array **T**, the resources used for the management of such data are negligible. Figure 7 shows two search-patterns examples that have been generated by the HS-BM approach. Such patterns exhibit the evaluated search-locations (rule 1 and 2) in white-cells, whereas the minimum location is marked in black. Grey-cells represent cells that have been estimated (rule 3) or not visited during the optimization process.

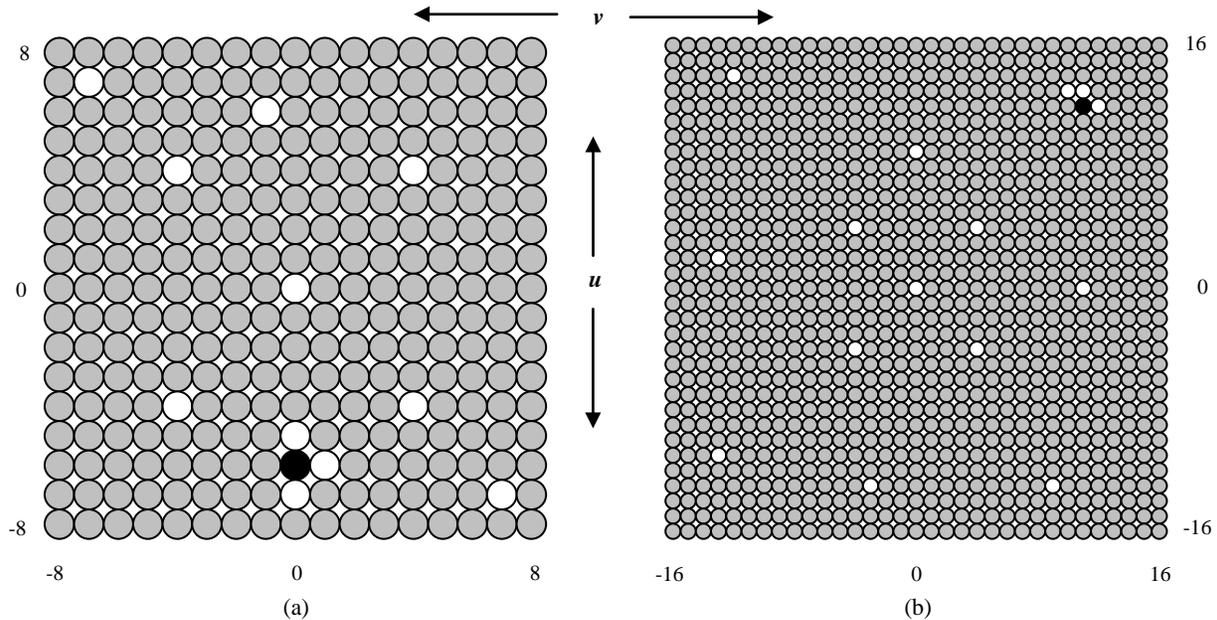

**Fig. 7.** Search-patterns generated by the HS-BM algorithm. (a) Search window pattern ±8 with solution $\hat{u}_{best}^1 = 0$ and $\hat{v}_{best}^1 = -6$. (b) Search window pattern ±16 with solution $\hat{u}_{best}^2 = 11$ and $\hat{v}_{best}^2 = 12$.

*5.4 Discussion on the accuracy of the fitness approximation strategy*

HS has been found to be capable of solving several practical optimization problems. A distinguishing feature of HS is about its operation with only a population of individuals. It uses multiple candidate solutions at each step. This requires the computation of the fitness function for each candidate at every iteration. The ability to locate the global optimum depends on sufficient exploration of the search space which requires the use of enough individuals. Under such circumstances, this work proposes to couple the HS method with a fitness approximation model in order to replace (when it is feasible) the use of an expensive fitness function to compute the quality of several individuals.

Similar to other EA approaches, HS maintains two different phases on its operation: exploitation and exploration [58]. Exploitation (local search) refers to the action of refining the best solutions found so far whereas exploration (global search) represents the process of generating semi-random individuals in order to capture information of unexplored areas. In spite of this, the optimization process is guided by the best individuals seen-so-far [59]. They are selected more frequently and thereby modified or combined by the evolutionary operators in order to generate new promising individuals.

Therefore, the main concern in using fitness approximation models, is to accurately calculate the quality of those individuals which either hold great possibilities of grasping an excellent fitness value (individuals that are to close to one of the best individuals seen-so-far), or do not have reference about their possible fitness values (individuals located in unexplored areas) [60,61]. Most of the fitness approximation methods proposed in the literature [48]-[50] use interpolation models in order to compute the fitness value of new individuals. Since the estimated fitness value is approximated considering other individuals which might be located far away from the position to be calculated, it introduces big errors that harshly affects the optimization procedure [45]. Different to such methods, in our approach, the fitness values are calculated using three different rules which promote the evaluation of individuals that require particular accuracy (Rule 1 and Rule 2). On the other hand, the strategy estimates those individuals which according to the evidence known so-far (elements contained in the array **T**) represent unacceptable solutions (bad fitness values). Such individuals do not play an important roll in the optimization process, therefore their accuracy is not considered critical [46,62].

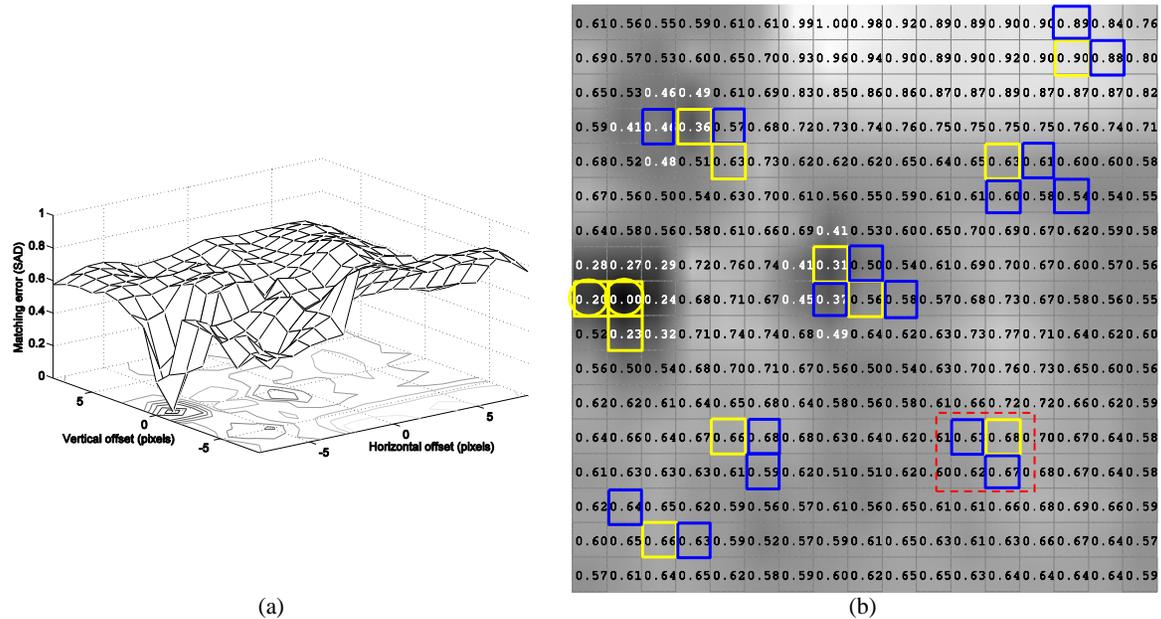

(a)            (b)

**Fig. 8.** Example of the optimization procedure: (a) Error landscape (SAD values) in a 3-D view. (b) Search positions calculated by the fitness approximation strategy over the SAD values which are computed for all elements of the search window of size ±8.

It is important to emphasize that the proposed fitness approximation strategy has been designed considering some of the BM process particularities. Error landscapes in BM, due to the continuity principle [17,20,63] of video sequences, present the following particularity: the closer neighbors to one global/local minimum (a motion vector with a low SAD value) decrement their SAD value as they approach to it. Such behavior is valid even in the most complex movement types. Under such circumstances, when it is necessary calculate the fitness value of a search position which is close to one of the search position previously visited (according to array **T**) and whose fitness value was unacceptable, its fitness value is estimated according to Rule 3. This decision is taken considering that there is a strong evidence to consider such position as a bad individual from which it is no necessary to get a good accuracy level.

Fig. 8 presents the optimization procedure achieved by the combination between HS and the proposed fitness approximation strategy over a complex movement case. The example illustrates the fitness strategy operation for a complex movement considering a search window of ±8. Fig. 8a shows the error landscape (SAD values) in a 3-D view, whereas Fig. 8b depicts the search positions calculated by the fitness approximation strategy over the SAD values that are computed for all elements of the search window as reference (for the sake of representation, both Figures are normalized from 0 to 1). Yellow squares indicate evaluated search positions whereas blue squares represent the estimated ones. Since random numbers are involved by HS in the generation of new individuals, they may encounter same positions (repetition) that other individuals have visited in previous iterations. Circles represent search positions that have been selected several times during the optimization procedure. The problem of accuracy, in the estimation process, can also be appreciated through a close analysis from the red dashed square of Fig. 8b. As the blue squares represent the estimated search positions according to the Rule 3, their fitness values are both assigned to 0.68 substituting their actual value of 0.63 and 0.67 respectively. Thus, considering that such individuals present an unacceptable solution (according to the elements stored in the array **T**), the differences in the fitness value are negligible for the optimization process. From Fig. 8b, it can be seen that although the fitness function considers 30 individuals only 12 are actually evaluated by the fitness function (note that circle positions represent multiple evaluations).

## 6. Experimental results

*6.1 HS-BM results*

This section presents the results of comparing the proposed HS-BM algorithm with other existing fast BMAs. The simulations have been performed over the luminance component of popular video sequences that are listed in Table 1. Such sequences consist of different degrees and types of motion including QCIF (176x144), CIF (352x288) and SIF (352x240) respectively. The first four sequences are *Container*, *Carphone*, *Foreman* and *Akiyo* in QCIF format. The next two sequences are *Stefan* in CIF format and *Football* in SIF format. Among such sequences, *Container* has gentle, smooth and low motion changes and consists mainly of stationary and quasi-stationary blocks. *Carphone*, *Foreman* and *Akiyo* have moderately complex motion getting a ''medium'' category regarding its motion content. Rigorous motion which is based on camera panning with translation and complex motion content can be found in the sequences of *Stefan* and *Football*. Figure 9 shows a sample frame from each sequence.

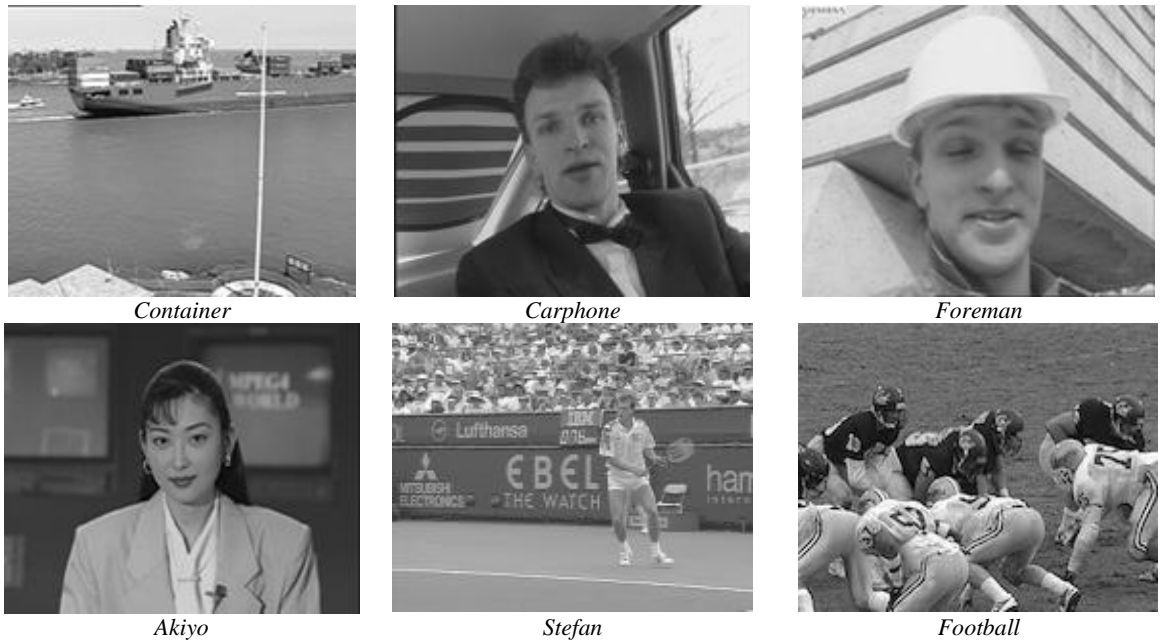

| *Container* | *Carphone* | *Foreman* |
| *Akiyo* | *Stefan* | *Football* |

**Fig. 9.** Test video sequences.

Each picture frame is partitioned into macro-blocks with the sizes of 16x16 (*N*=16) pixels for motion estimation, where the maximum displacement within the search range *W* is of ±8 pixels in both horizontal and vertical directions for the sequences *Container*, *Carphone*, *Foreman*, *Akiyo* and Stefan. The sequence *Football* has been simulated with a window size *W* of ±16, which requires a greater number of iterations (8 iterations) by the HS-BM method.

In order to compare the performance of the HS-BM approach, different search algorithms such as FSA, TSS [12], 4SS [15], NTSS [13], BBGD [19], DS [16], NE [20], ND [22], LWG [29], GFSS [30] and PSO-BM [31] have been all implemented in our simulations. For comparison purposes, all six video sequences in Fig. 8 have been all used. All simulations are performed on a Pentium IV 3.2 GHz PC with 1GB of memory.

In the comparison, two relevant performance indexes have been considered: the distortion performance and the search efficiency.

**Table 1.** Test sequences used in the comparison test.

| Sequence | Format | Total frames | Motion type |
|---|---|---|---|
| *Container* | QCIF(176x144) | 299 | Low |
| *Carphone* | QCIF(176x144) | 381 | Medium |
| *Foreman* | QCIF(352x288) | 398 | Medium |
| *Akiyo* | QCIF(352x288) | 211 | Medium |
| *Stefan* | CIF(352x288) | 89 | High |
| *Football* | SIF(352x240) | 300 | High |

Distortion performance

First, all algorithms are compared in terms of their distortion performance which is characterized by the Peak-Signal-to-Noise-Ratio (PSNR) value. Such value indicates the reconstruction quality when motion vectors, which are computed through a BM approach, are used. In PSNR, the signal is the original data frames whereas the noise is the error introduced by the calculated motion vectors. The PSNR is thus defined as:

$$\text{PSNR} = 10 \cdot \log_{10}\left(\frac{255^2}{MSE}\right) \tag{7}$$

where *MSE* is the mean square between the original frames and those compensated by the motion vectors. Additionally, as an alternative performance index, it is used in the comparison the PSNR degradation ratio ($D_{\text{PSNR}}$). This ratio expresses in percentage (%) the level of mismatch between the PSNR of a BM approach and the PSNR of the FSA which is considered as reference. Thus the $D_{\text{PSNR}}$ is defined as

$$D_{\text{PSNR}} = -\left(\frac{\text{PSNR}_{\text{FSA}} - \text{PSNR}_{\text{BM}}}{\text{PSNR}_{\text{FSA}}}\right) \cdot 100\% \tag{8}$$

Table 2 shows the comparison of the PSNR values and the PSNR degradation ratios ($D_{\text{PSNR}}$) among the BM algorithms. The experiment considers the six image sequences presented in Fig. 8. As it can be seen, in the case of the slow-moving sequence *Container*, the PSNR values (the $D_{\text{PSNR}}$ ratios) of all BM algorithms are similar. For the medium motion content sequences such as *Carphone*, *Foreman* and *Akiyo*, the approaches consistent of fixed patterns (TSS, 4SS and NTSS) exhibit the worst PSNR value (high $D_{\text{PSNR}}$ ratio) except for the DS algorithm. On the other hand, BM methods that use evolutionary algorithms (LWG, GFSS, PSO-BM and HS-BM) present the lowest $D_{\text{PSNR}}$ ratio, only one step under the FSA approach which is considered as reference. Finally, approaches based on the error-function minimization (BBGD and NE) and pixel-decimation (ND), show an acceptable performance. For the high motion sequence of *Stefan*, since the motion content of these sequences is complex producing error surfaces with more than one minimum, the performance, in general, becomes worst for most of the algorithms especially for those based on fixed patterns. In the sequence *Football,* which has been simulated with a window size of ±16, the algorithms based on the evolutionary algorithms present the best PSNR values. Such performance is because evolutionary methods adapt better to complex optimization problems where the search area and the number of local minima increase. As a summary of the distortion performance, the last column of Table 2 presents the average PSNR degradation ratio ($D_{PSNR}$) obtained for all sequences. According to such values, the proposed HS-BM method is superior to any other approach. Due to the computation complexity, the FSA is considered just as a reference. The best entries are bold-cased in Table 2.

**Table 2.** PSNR values and $D_{PSNR}$ comparison of the BM methods

| Algorithm | Container $W=\pm 8$ | | Carphone $W=\pm 8$ | | Foreman $W=\pm 8$ | | Akiyo $W=\pm 8$ | | Stefan $W=\pm 8$ | | Football $W=\pm 16$ | | Total Average ($D_{PSNR}$) |
|---|---|---|---|---|---|---|---|---|---|---|---|---|---|
| | PSNR | $D_{PSNR}$ | PSNR | $D_{PSNR}$ | PSNR | $D_{PSNR}$ | PSNR | $D_{PSNR}$ | PSNR | $D_{PSNR}$ | PSNR | $D_{PSNR}$ | |
| FSA | 43.18 | 0 | 31.51 | 0 | 31.69 | 0 | 29.07 | 0 | 25.95 | 0 | 23.07 | 0 | 0 |
| TSS | 43.10 | -0.20 | 30.27 | -3.92 | 29.37 | -7.32 | 26.21 | -9.84 | 21.14 | -18.52 | 20.03 | -13.17 | -8.82 |
| 4SS | 43.12 | -0.15 | 30.24 | -4.01 | 29.34 | -7.44 | 26.21 | -9.84 | 21.41 | -17.48 | 20.10 | -12.87 | -8.63 |
| NTSS | 43.12 | -0.15 | 30.35 | -3.67 | 30.56 | -3.57 | 27.12 | -6.71 | 22.52 | -13.20 | 20.21 | -12.39 | -6.61 |
| BBGD | 43.14 | -0.11 | 31.30 | -0.67 | 31.00 | -2.19 | 28.10 | -3.33 | 25.17 | -3.01 | 22.03 | -4.33 | -2.27 |
| DS | 43.13 | -0.13 | 31.26 | -0.79 | 31.19 | -1.59 | 28.00 | -3.70 | 24.98 | -3.73 | 22.35 | -3.12 | -2.17 |
| NE | 43.15 | -0.08 | 31.36 | -0.47 | 31.23 | -1.47 | 28.53 | -2.69 | 25.22 | -2.81 | 22.66 | -1.77 | -1.54 |
| ND | 43.15 | -0.08 | 31.35 | -0.50 | 31.20 | -1.54 | 28.32 | -2.56 | 25.21 | -2.86 | 22.60 | -2.03 | -1.59 |
| LWG | 43.16 | -0.06 | 31.40 | -0.36 | 31.31 | -1.21 | 28.71 | -1.22 | 25.41 | -2.09 | 22.90 | -0.73 | -0.95 |
| GFSS | 43.15 | -0.06 | 31.38 | -0.40 | 31.29 | -1.26 | 28.69 | -1.28 | 25.34 | -2.36 | 22.92 | -0.65 | -1.01 |
| PSO-BM | 43.15 | -0.07 | 31.39 | -0.38 | 31.27 | -1.34 | 28.65 | 1.42 | 25.39 | -2.15 | 22.88 | -0.82 | -1.03 |
| HS-BM | 43.16 | **-0.03** | 31.49 | **-0.03** | 31.63 | **-0.21** | 29.01 | **-0.18** | 25.89 | **-0.20** | 23.01 | **-0.20** | **-0.18** |

Search efficiency

The search efficiency is used in this section as a measurement of computational complexity. The search efficiency is calculated by counting the average number of search points (or the average number of SAD computations) for the MV estimation. In Table 3, the search efficiency is compared, where the best entries are bold-cased. Just above FSA, some evolutionary algorithms such as LWG, GFSS and PSO-BM hold the highest number of search points per block. On the contrary, the proposed HS-BM algorithm can be considered as a fast approach as it maintains a similar performance to DS. From data shown in Table 3, the average number of search locations, corresponding to the HS-BM method, represents the number of SAD evaluations (the number of SAD estimations are not considered whatsoever). Additionally, the last two columns of Table 3 present the number of search locations that have been averaged (over the six sequences) and their performance rank. According to these values, the proposed HS-BM method is ranked in the first place. The average number of search points visited by the HS-BM algorithm ranges from 9.2 to 17.3, representing the 4% and the 7.4% respectively in comparison to the FSA method. Such results demonstrate that our approach can significantly reduce the number of search points. Hence, the HS-BM algorithm proposed in this paper is comparable to the fastest algorithms and delivers a similar precision to the FSA approach.

**Table 3.** Averaged number of visited search points per block for all ten BM methods.

| Algorithm | Container $W=\pm 8$ | Carphone $W=\pm 8$ | Foreman $W=\pm 8$ | Akiyo $W=\pm 8$ | Stefan $W=\pm 8$ | Football $W=\pm 16$ | Total Average | Rank |
|---|---|---|---|---|---|---|---|---|
| FSA | 289 | 289 | 289 | 289 | 289 | 1089 | 422.3 | 12 |
| TSS | 25 | 25 | 25 | 25 | 25 | 25 | 25 | 8 |
| 4SS | 19 | 25.5 | 24.8 | 27.3 | 25.3 | 25.6 | 24.58 | 7 |
| NTSS | 17.2 | 21.8 | 22.1 | 23.5 | 25.4 | 26.5 | 22.75 | 6 |
| BBGD | 9.1 | 14.5 | 14.5 | 13.2 | 17.2 | 22.3 | 15.13 | 3 |
| DS | 7.5 | 12.5 | 13.4 | 11.8 | **15.2** | 17.8 | 13.15 | 2 |
| NE | 11.7 | 13.8 | 14.2 | 14.5 | 19.2 | 24.2 | 16.36 | 5 |
| ND | 10.8 | 13.4 | 13.8 | 14.1 | 18.4 | 25.1 | 16.01 | 4 |
| LWG | 75 | 75 | 75 | 75 | 75 | 75 | 75 | 11 |
| GFSS | 60 | 60 | 60 | 60 | 60 | 60 | 60 | 10 |
| PSO-BM | 32.5 | 48.5 | 48.1 | 48.5 | 52.2 | 52.2 | 47 | 9 |
| HS-BM | **8.0** | **12.2** | **11.2** | **11.5** | 17.1 | **15.2** | **12.50** | **1** |

*6.2 Results on H.264*

Other set of experiments have been performed in JM-12.2 [64] of H.264/AVC reference software. In the simulations, we compare FS, DS [16], EPZS [24], TSS [12], 4SS [15], NTSS [13], BBGD [19] and the proposed HS-BM algorithm in terms of coding efficiency and computational complexity.

For encoding purposes JM-12.2 Main Encoder Profile has been used. For each test sequence only the first frame has been coded as I frame and the remaining frames are coded as P frames. Only one reference frame has been used. Each pixel in the image sequences is uniformly quantized to 8 bits. Sum of absolute difference (SAD) distortion function is used as the block distortion measure (BDM). Image formats used are QCIF, CIF and SIF meanwhile sequences are tested at 30 fps (frames per second). The simulation platform in our experiments is a PC with Intel Pentium IV 2.66 GHz CPU.

The test sequences used for our experiments are *Container*, *Akiyo* and *Football*. These sequences exhibit a variety of motion that is generally encountered in real video. For the sequences *Container* and *Akiyo* a search window of ±8 is selected meanwhile for the football sequence a search window of ±16 is considered. The group of experiments has been performed over such sequences at four different quantization parameters (QP=28,32,36,40) in order to test the algorithms at different transmission conditions.

*a) Coding efficiency*

In the first experiment, the performance of the proposed algorithm is compared to other BM algorithms regarding the coding efficiency. Two different performance indexes are used for evaluating the coding quality: the PSNR Gain and the increasing of the Bit Rate. In order to comparatively assess the results, two additional indexes, called PSNR loss and Bit Rate Incr., relate the performance of each method with the FSA performance as a reference. Such indexes are calculated as follows:

$$\text{PSNR loss} = \text{PSNR FSA} - \text{PSNR algorithm} \quad (9)$$

$$\text{Bit Rate Incr.} = \left( \frac{\text{Bit Rate algorithm} - \text{Bit Rate FSA}}{\text{Bit Rate FSA}} \right) \cdot 100 \quad (10)$$

Tables 4–6 show a coding efficiency comparison among BM algorithms. It is observed, from experimental results, that the proposed HS-BM algorithm holds an effective coding quality because the loss in terms of PSNR and the increase of the Bit rate are low with an average of 1.6dB and -0.04%, respectively. Such coding performance is similar to the one produced by the EPZS method whereas it is much better than the obtained by other BM algorithms which posses the worst coding quality.

**Table 4.** Coding efficiency results for the *container* sequence, considering a window size $W$ of $\pm 8$.

| BM | Coding efficiency | | | | Computational complexity | | |
|---|---|---|---|---|---|---|---|
|  | PSNR | Bit-rate (Kbits/s) | PNSR loss (dB) | Bit-rate increase (%) | ACT (ms) | IN | CM (Bytes) |
| FSA | 36.06 | 41.4 | - | - | 133.2 | 122 | 3072 |
| DS | 36.04 | 43.4 | 0.02 | 4.83 | 6.33 | 138 | 420 |
| EPZS | 36.04 | 41.3 | 0.02 | -0.20 | 19.5 | 621 | 8972 |
| TSS | 34.01 | 45.2 | 2.05 | 9.17 | 2.1 | 100 | 180 |
| 4SS | 35.22 | 44.7 | 0.84 | 7.97 | 2.8 | 100 | 204 |
| NTSS | 35.76 | 44.3 | 0.30 | 7.00 | 3.7 | 110 | 256 |
| BBGD | 35.98 | 42.1 | 0.08 | 1.70 | 9.1 | 256 | 1024 |
| HS-BM | 36.04 | 41.5 | 0.02 | 0.20 | 3.8 | 189 | 784 |

**Table 5.** Simulation results for the *Akiyo* sequence, considering a window size $W$ of $\pm 8$.

| BM | Coding efficiency | | | | Computational complexity | | |
|---|---|---|---|---|---|---|---|
|  | PSNR | Bit-rate (Kbits/s) | PNSR loss (dB) | Bit-rate increase (%) | ACT (ms) | IN | CM (Bytes) |
| FSA | 38.19 | 25.6 | - | - | 133.2 | 122 | 3072 |
| DS | 38.11 | 25.9 | 0.08 | 1.20 | 7.45 | 138 | 420 |
| EPZS | 38.19 | 25.3 | - | -1.20 | 22.1 | 621 | 8972 |
| TSS | 30.32 | 29.3 | 7.87 | 14.45 | 2.1 | 100 | 180 |
| 4SS | 32.42 | 28.4 | 5.77 | 10.93 | 2.8 | 100 | 204 |
| NTSS | 33.57 | 27.2 | 4.62 | 6.25 | 3.7 | 110 | 256 |
| BBGD | 35.21 | 26.8 | 2.98 | 4.68 | 10.1 | 256 | 1024 |
| HS-BM | 38.17 | 25.5 | 0.02 | -0.40 | 3.9 | 189 | 784 |

**Table 6.** Simulation results for the *Football* sequence, considering a window size $W$ of $\pm 16$.

| BM | Coding efficiency | | | | Computational complexity | | |
|---|---|---|---|---|---|---|---|
|  | PSNR | Bit-rate (Kbits/s) | PNSR loss (dB) | Bit-rate increase (%) | ACT (ms) | IN | CM (Bytes) |
| FSA | 34.74 | 98.85 | - | - | 245.7 | 122 | 12288 |
| DS | 32.22 | 99.89 | 2.52 | 1.02 | 10.36 | 144 | 600 |
| EPZS | 34.72 | 98.81 | 0.02 | -0.04 | 26.8 | 678 | 20256 |
| TSS | 27.12 | 106.42 | 7.62 | 7.65 | 2.9 | 113 | 180 |
| 4SS | 27.91 | 105.29 | 6.83 | 6.51 | 3.1 | 113 | 204 |
| NTSS | 29.11 | 104.87 | 5.63 | 6.09 | 4.2 | 113 | 256 |
| BBGD | 29.76 | 103.96 | 4.98 | 5.19 | 16.41 | 268 | 2048 |
| HS-BM | 34.73 | 98.91 | 0.01 | 0.06 | 4.1 | 201 | 1024 |

*b) Computational complexity*

In the second experiment, we have compared the performance of the proposed algorithm to other BM algorithms in terms of computational overhead. As the JM-12.2 platform allows to simulate BM algorithms in real time conditions, we have used such results in order to evaluate their performances.

Three different performance indexes are used for evaluating the computational complexity; they are the Averaged Coding Time (ACT), Instruction Number (IN) and Consumed Memory (CM). The ACT is the averaged time employed to codify a complete frame (the averaged time consumed after finding all the corresponding motion vectors for a frame). IN represents the number of instructions used to implement each algorithm in the JM-12.2 profile. CM considers the memory size used by the JM-12.2 platform in order to manage the data that are employed by each BM algorithm.

Tables 4–6 show the computational complexity comparison among the BM algorithms. It is observed from the experimental results that the proposed HS-BM algorithm possesses a competitive ACT value (from 3.8 to 4.1 milliseconds) in comparison to other BM algorithms. This fact reflexes that although the cost of applying the fitness approximation strategy represents an overhead that is not required in most fast BM methods, such overhead is negligible in comparison to the cost of the number of fitness evaluations which have been saved. The ACT values, presented by the HS-BM, are lightly superior to those produced by the fast BM methods (TSS, 4SS and NTSS) whereas it is much better than those generated by the EPZS algorithm which posseses the worst computational performance. On the other hand, the resources (in terms of number of instructions IN and required memory CM) needed by the proposed approach are considered as standard in software and hardware architectures.

## 7. Conclusions

In this paper, a new BM algorithm that combines HS with a fitness approximation model is proposed. The approach uses as potential solutions the motion vectors belonging to the search window. A fitness function evaluates the matching quality of each motion vector candidate. To save computational time, the approach incorporates a fitness calculation strategy to decide which motion vectors can be estimated or actually evaluated. Guided by the values given by such fitness calculation strategy, the set of motion vectors are evolved using the HS operators so the best possible motion vector can be identify.

Since the proposed algorithm does not consider any fixed search pattern during the BM process or any other movement assumption, a high probability for finding the true minimum (accurate motion vector) is expected regardless of the movement complexity contained in the sequence. Therefore, the chance of being trapped into a local minimum is reduced in comparison to other BM algorithms.

The performance of HS-BM has been compared to other existing BM algorithms by considering different sequences which present a great variety of formats and movement types. Experimental results demonstrate that the proposed algorithm maintains the best balance between coding efficiency and computational complexity.

Although the experimental results indicate that the HS-BM method can yield better results on complicated sequences, it should be noticed that the aim of our paper is not intended to beat all the BM methods which have been proposed earlier, but to show that the fitness approximation can effectively serve as an attractive alternative to evolutionary algorithms for solving complex optimization problems, yet demanding fewer function evaluations.

## References

[1] Giansalvo Cirrincione and Maurizio Cirrincione. A Novel Self-Organizing Neural Network for Motion Segmentation, Applied Intelligence, 18(1), (2003), 27-35.

[2] Lon Risinger and Khosrow Kaikhah. Motion detection and object tracking with discrete leaky integrate-and-fire neurons, Applied Intelligence, 29(3), (2008), 248-262.

[3] Ali Bohlooli and Kamal Jamshidi. A GPS-free method for vehicle future movement directions prediction using SOM for VANET, Applied Intelligence, 36(3), (2012), 685-697.

Please cite this article as:
**Cuevas, E. Block-matching algorithm based on harmony search optimization for motion estimation,** *Applied Intelligence*, 39 (1), (2013), pp. 165-183[4] Jeong-Gwan Kang, Sunhyo Kim, Su-Yong An and Se-Young Oh. A new approach to simultaneous localization and map building with implicit model learning using neuro evolutionary optimization, Applied Intelligence, 36(1), (2012), 242-269.

[5] Dimitrios Tzovaras, Ioannis Kompatsiaris, Michael G. Strintzis. 3D object articulation and motion estimation in model-based stereoscopic videoconference image sequence analysis and coding. Signal Processing: Image Communication, 14(10), 1999, 817-840.

[6] Barron, J.L., Fleet, D.J., Beauchemin, S.S., 1994. Performance of optical flow techniques. Int. J. Comput. Vision 12 (1), 43–77.

[7] J. Skowronski. Pel recursive motion estimation and compensation in subbands. Signal Processing: Image Communication 14, (1999), 389-396.

[8] Huang, T., Chen, C., Tsai, C., Shen, C., Chen, L. Survey on Block Matching Motion Estimation Algorithms and Architectures with New Results. Journal of VLSI Signal Processing 42, 297–320, 2006.

[9] MPEG4, Information Technology Coding of Audio Visual Objects Part 2: Visual, JTC1/SC29/WG11, ISO/IEC14469-2(MPEG-4Visual), 2000.

[10] H.264, Joint Video Team (JVT) of ITU-T and ISO/IEC JTC1, Geneva, JVT ofISO/IEC MPEG and ITU-T VCEG, JVT-g050r1, Draft ITU-TRec. and Final Draft International Standard of Joint Video Specification (ITU-T Rec.H.264-ISO/IEC14496-10AVC), 2003.

[11] J. R. Jain and A. K. Jain, Displacement measurement and its application in interframe image coding, IEEE Trans. Commun., vol. COM-29, pp. 1799–1808, Dec. 1981.

[12] H.-M. Jong, L.-G. Chen, and T.-D. Chiueh, "Accuracy improvement and cost reduction of 3-step search block matching algorithm for video coding," IEEE Trans. Circuits Syst. Video Technol., vol. 4, pp. 88–90, Feb. 1994.

[13] Renxiang Li, Bing Zeng, and Ming L. Liou, "A New Three-Step Search Algorithm for Block Motion Estimation", IEEE Trans. Circuits And Systems For Video Technology, vol 4., no. 4, pp. 438-442, August 1994.

[14] Jianhua Lu, and Ming L. Liou, "A Simple and Efficent Search Algorithm for Block-Matching Motion Estimation", IEEE Trans. Circuits And Systems For Video Technology, vol 7, no. 2, pp. 429-433, April 1997

[15] Lai-Man Po, and Wing-Chung Ma, "A Novel Four-Step Search Algorithm for Fast Block Motion Estimation", IEEE Trans. Circuits And Systems For Video Technology, vol 6, no. 3, pp. 313-317, June 1996.

[16] Shan Zhu, and Kai-Kuang Ma, " A New Diamond Search Algorithm for Fast Block-Matching Motion Estimation", IEEE Trans. Image Processing, vol 9, no. 2, pp. 287-290, February 2000.

[17] Yao Nie, and Kai-Kuang Ma, Adaptive Rood Pattern Search for Fast Block-Matching Motion Estimation, IEEE Trans. Image Processing, vol 11, no. 12, pp. 1442-1448, December 2002.

[18] Yi-Ching L., Jim L., Zuu-Chang H. Fast block matching using prediction and rejection criteria. Signal Processing, 89, (2009), pp 1115–1120.

[19] Liu, L., Feig, E. A block-based gradient descent search algorithm for block motion estimation in video coding, IEEE Trans. Circuits Syst. Video Technol., 6(4),(1996),419–422.22

[20] Saha, A., Mukherjee, J., Sural, S. A neighborhood elimination approach for block matching in motion estimation, Signal Process Image Commun, (2011), 26, 8–9, 2011, 438–454.

[21] K.H.K. Chow, M.L. Liou, Generic motion search algorithm for video compression, IEEE Trans. Circuits Syst. Video Technol. 3, (1993), 440–445.

[22] A. Saha , J. Mukherjee, S. Sural. New pixel-decimation patterns for block matching in motion estimation. Signal Processing: Image Communication 23 (2008)725–738.

[23] Y. Song, T. Ikenaga, S. Goto. Lossy Strict Multilevel Successive Elimination Algorithm for Fast Motion Estimation. IEICE Trans. Fundamentals E90(4), 2007, 764-770.

[24] A.M. Tourapis, Enhanced predictive zonal search for single and multiple frame motion estimation, in: Proceedings of Visual Communications and Image Processing, California, USA, January 2002, pp. 1069–1079.

[25] Chen, Z., Zhou, P., He, Y., Chen, Y., 2002. Fast Integer Pel and Fractional Pel Motion Estimation for JVT, December, 2002, ITU-T. Doc. #JVT-F-017.

[26] Humaira Nisar, Aamir Saeed Malik, Tae-Sun Choi. Content adaptive fast motion estimation based on spatio-temporal homogeneity analysis and motion classification. Pattern Recognition Letters 33 (2012) 52–61

[27] J.H. Holland, Adaptation in Natural and Artificial Systems, University of Michigan Press, Ann Arbor, MI, 1975.

[28] J. Kennedy, R.C. Eberhart, Particle swarm optimization, in: Proceedings of the 1995 IEEE International Conference on Neural Networks, vol. 4, 1995, pp. 1942–1948.

[29] Chun-Hung, L., Ja-Ling W. A Lightweight Genetic Block-Matching Algorithm for Video Coding. IEEE Transactions on Circuits and Systems for Video Technology, 8(4), (1998), 386-392.

[30] Wu, A., So, S. VLSI Implementation of Genetic Four-Step Search for Block Matching Algorithm. IEEE Transactions on Consumer Electronics, 49(4), (2003), 1474-1481.

[31] Yuan, X., Shen, X. Block Matching Algorithm Based on Particle Swarm Optimization. International Conference on Embedded Software and Systems (ICESS2008), 2008, Sichuan, China.

[32] Z.W. Geem, J.H. Kim, G.V. Loganathan, A new heuristic optimization algorithm: harmony search, Simulations 76 (2001) 60–68.

[33] M. Mahdavi, M. Fesanghary, E. Damangir, An improved harmony search algorithm for solving optimization problems, Appl. Math. Comput. 188 (2007) 1567–1579.

[34] M.G.H. Omran, M. Mahdavi, Global-best harmony search, Appl. Math. Comput. 198 (2008) 643–656.

[35] K.S. Lee, Z.W. Geem, A new meta-heuristic algorithm for continuous engineering optimization, harmony search theory and practice, Comput. Methods Appl. Mech. Eng. 194 (2005) 3902–3933.

[36] K.S. Lee, Z.W. Geem, S. H Lee, K.-W. Bae, The harmony search heuristic algorithm for discrete structural optimization, Eng. Optim. 37 (2005) 663–684.

[37] J.H. Kim, Z.W. Geem, E.S. Kim, Parameter estimation of the nonlinear Muskingum model using harmony search, J. Am. Water Resour. Assoc. 37 (2001) 1131–1138.

[38] Z.W. Geem, Optimal cost design of water distribution networks using harmony search, Eng. Optim. 38 (2006) 259–280.

[39] K.S. Lee, Z.W. Geem, A new structural optimization method based on the harmony search algorithm, Comput. Struct. 82 (2004) 781–798.

[40] T.M. Ayvaz, Simultaneous determination of aquifer parameters and zone structures with fuzzy c-means clustering and meta-heuristic harmony search algorithm, Adv. Water Resour. 30 (2007) 2326–2338.

[41] Z.W. Geem, K.S. Lee, Y.J. Park, Application of harmony search to vehicle routing, Am. J. Appl. Sci. 2 (2005) 1552–1557.

[42] Z.W. Geem, Novel derivative of harmony search algorithm for discrete design variables, Appl. Math. Comput. 199 (1) (2008) 223–230.

[43] Erik Cuevas, Noé Ortega-Sánchez, Daniel Zaldivar and Marco Pérez-Cisneros. Circle Detection by Harmony Search Optimization. Journal of Intelligent & Robotic Systems, 2012, 66(3), 359-376.

[44] Jin, Y. Comprehensive survey of fitness approximation in evolutionary computation. Soft Computing, 9, (2005), 3–12.

[45] Yaochu Jin. Surrogate-assisted evolutionary computation: Recent advances and future challenges. Swarm and Evolutionary Computation, 1, (2011), 61–70.

[46] J. Branke, C. Schmidt. Faster convergence by means of fitness estimation. Soft Computing 9, (2005), 13–20.

[47] Zhou, Z., Ong, Y., Nguyen, M., Lim, D. A Study on Polynomial Regression and Gaussian Process Global Surrogate Model in Hierarchical Surrogate-Assisted Evolutionary Algorithm, IEEE Congress on Evolutionary Computation (ECiDUE'05), Edinburgh, United Kingdom, September 2-5, 2005.

[48] Ratle, A. Kriging as a surrogate fitness landscape in evolutionary optimization. Artificial Intelligence for Engineering Design, Analysis and Manufacturing, 15, (2001), 37–49.

[49] Lim, D., Jin, Y., Ong, Y., Sendhoff, B. Generalizing Surrogate-assisted Evolutionary Computation, IEEE Transactions on Evolutionary Computation, 14( 3), (2010),  329-355.

[50] Ong, Y., Lum, K., Nair, P. Evolutionary Algorithm with Hermite Radial Basis Function Interpolants for Computationally Expensive Adjoint Solvers, Computational Optimization and Applications, 39(1), (2008), 97-119.

[51] Luoa, C.,  Shao-Liang, Z., Wanga, C., Jiang, Z. A metamodel-assisted evolutionary algorithm for expensive optimization. Journal of Computational and Applied Mathematics, doi:10.1016/j.cam.2011.05.047, (2011).

[52] Goldberg, D. E. Genetic algorithms in search, optimization and machine learning. Menlo Park, (1989) CA: Addison-Wesley Professional.

[53] Li, X., Xiao, N.,  b, Claramunt, C., Lin, H. Initialization strategies to enhancing the performance of genetic algorithms for the p-median problem, Computers & Industrial Engineering, (2011), doi:10.1016/j.cie.2011.06.015.

[54] Xiao, N. A unified conceptual framework for geographical optimization using evolutionary algorithms. Annals of the Association of American Geographers, 98, (2008), 795–817.

[55] Sang-Moon Soak and Sang-Wook Lee. A memetic algorithm for the quadratic multiple container packing problem, Applied Intelligence, 36(1), (2012), 119-135.

[56] Cristobal Luque, Jose M. Valls and Pedro Isasi. Time series prediction evolving Voronoi regions, Applied Intelligence, 34(1), (2011), 116-126.

[57] Elizabeth Montero, María-Cristina Riff. On-the-fly calibrating strategies for evolutionary algorithms, Information Sciences, 181, (2011), 552–566.

[58] K.C. Tan, S.C. Chiam, A.A. Mamun, C.K. Goh. Balancing exploration and exploitation with adaptive variation for evolutionary multi-objective optimization, European Journal of Operational Research, 197( 2), (2009), 701-713.

[59] Pan Wang, Jianjian Zhang, Li Xub, Hong Wang, Shan Feng, Haoshen Zhu. How to measure adaptation complexity in evolvable systems – A new synthetic approach of constructing fitness functions, Expert Systems with Applications, 38, (2011), 10414–10419.

[60] Yoel Tenne. A computational intelligence algorithm for expensive engineering optimization problems, Engineering Applications of Artificial Intelligence, 25(5), (2012), 1009-1021.

[61] Dirk Büche, Nicol N. Schraudolph, Petros Koumoutsakos. Accelerating Evolutionary Algorithms With Gaussian Process Fitness Function Models, IEEE Transactions on systems, man, and cybernetics—part c: applications and reviews, 35(2), (2005), 183-194.

[62] K.C. Giannakoglou, D.I. Papadimitriou, I.C. Kampolis. Aerodynamic shape design using evolutionary algorithms and new gradient-assisted metamodels, Comput. Methods Appl. Mech. Engrg. 195, (2006), 6312–6329.

[63] Shen-Chuan Tai, Ying-Ru Chen, Yu-Hung Chen. Small-diamond-based search algorithm for fast block motion estimation, Signal Processing: Image Communication 22, (2007), 877–890.

[64] Joint Video Team Reference Software, 2007. Version 12.2 (JM12.2). <http://iphome.hhi.de/suehring/tml/download/>.